\documentclass[lettersize,journal]{IEEEtran}
\usepackage[justification=centering]{caption}
\usepackage{amsmath,amsfonts}
\usepackage{algorithmic}
\usepackage{array}
\usepackage[caption=false,font=normalsize,labelfont=sf,textfont=sf]{subfig}
\usepackage{textcomp}
\usepackage{stfloats}
\usepackage{url}
\usepackage{verbatim}
\usepackage{graphicx}
\def\BibTeX{{\rm B\kern-.05em{\sc i\kern-.025em b}\kern-.08em
    T\kern-.1667em\lower.7ex\hbox{E}\kern-.125emX}}
\usepackage{balance}
\usepackage{cite}
\begin{document}
\captionsetup[figure]{labelfont={bf},name={Fig.},labelsep=period}
\captionsetup[table]{labelfont={bf},name={Table.},labelsep=period}
\title{Adding New Categories in Object Detection Using Few-Shot Copy-Paste}
\author{Boyang Deng, Meiyan Lin, Shoulun Long}


\maketitle

\begin{abstract}
Developing data-efficient instance detection models that can handle rare object categories remains a key challenge in computer vision. However, existing research often overlooks data collection strategies and evaluation metrics tailored to real-world scenarios involving neural networks. In this study, we systematically investigate data collection and augmentation techniques focused on object occlusion, aiming to mimic occlusion relationships observed in practical applications. Surprisingly, we find that even a simple occlusion mechanism is sufficient to achieve strong performance when introducing new object categories. Notably, by adding just 15 images of a new category to a large-scale training dataset containing over half a million images across hundreds of categories, the model achieves 95\% accuracy on an unseen test set with thousands of instances of the new category.
\end{abstract}


\section{Introduction}
Object detection is a fundamental task in computer vision with numerous real-world applications. However, state-of-the-art object detection models based on convolutional neural networks are typically data-hungry \cite{tian2019fcos}. Annotating large-scale datasets for object detection is both expensive and time-consuming. For instance, in our Smart Shelf dataset, it takes four workers approximately one hour to annotate just 3,000 object bounding boxes. This highlights the urgent need to develop methods that enhance the data efficiency of modern object detection models.

Many studies have attempted to boost detection performance through architectural innovations \cite{zhang2020bridging, rong2020solution, tian2019fcos}, such methods often introduce trade-offs, including increased inference time or added model complexity. In contrast, our focus is on developing generalizable strategies that enhance model performance through data augmentation techniques, as demonstrated in \cite{zhang2019bag}. We explore strategies for efficiently adding new categories to an existing dataset with minimal effort in image collection and annotation. One promising approach is few-shot learning with real images, which leverages a limited number of real images for the new categories while still aiming to maintain high detection accuracy.

We propose that effective management of data collection and augmentation is a direct and impactful way to enhance the data efficiency of object detection models. Training detection networks on diverse image distributions has shown notable benefits \cite{he2019rethinking}, and incorporating object occlusion can further enrich training data with challenging scenarios \cite{kuznetsova2020open}. In the data collection phase, natural occlusions are captured using real objects, while in the augmentation phase, occlusions are synthetically generated by overlaying extracted bounding boxes onto target objects.

Although bounding box annotations are significantly faster to obtain than segmentation masks, they may include partial background, leading to inconsistencies when pasted onto new images. This makes occlusion-based augmentation using bounding boxes less optimal than segmentation-based methods. Nevertheless, our experiments demonstrate that, with careful design, occlusion augmentation using bounding boxes can still yield substantial improvements in detection accuracy.

Inspired by recent data augmentation techniques \cite{zhang2017mixup, ghiasi2021simple}, we propose a new copy-paste-based method for training object detection networks using only bounding box annotations. Our approach aligns with the incremental learning concept introduced in \cite{shmelkov2017incremental}, aiming to efficiently incorporate new categories. However, in contrast to synthetic-only approaches \cite{hinterstoisser2019annotation}, we find that relying solely on synthetic data yields suboptimal results in real-world scenarios.

\section{Method}

The core idea of this study is to simulate object occlusions as they occur in real-world scenarios during training dataset construction. This approach enables the creation of diverse and combinatorial occlusion relationships with various possibilities, including:
\begin{enumerate} \item Selecting multiple objects that partially occlude one another;
\item Defining the occlusion relationships among these objects;
\item Determining object placements and camera viewpoints to capture the intended scene. \end{enumerate}

Our Copy-Paste-based data generation method introduces varying levels of occlusion to simulate realistic object interactions. We hypothesize that the occlusion relationship between objects is a critical factor for neural network learning, especially when training with a small number of annotated examples from a new category. By exposing the model to partial views of target objects, the method encourages robust feature learning under occlusion.

Experimental results suggest that the structure of occlusion, including its severity, viewpoint, and visible regions, is more influential than the specific categories of the occluding objects. This indicates that replicating realistic occlusion patterns can enable effective learning even with limited annotated data. We also find that annotating only the visible portions of objects, while ignoring occluded regions, leads to faster convergence and improved detection accuracy.

\subsection{Objects for occlusion}

In real-world object occlusion relationships, small objects typically occlude only partial portions of larger objects, while large objects can obscure significant portions of smaller objects. These occlusion relationships follow a relatively fixed distribution in natural settings, which provides an opportunity to replicate this distribution in our synthetic dataset by targeting important sample points.

For instance, consider a target object A from category X. If, in a given occlusion distribution, A’s bottom 50\% is occluded by an object B from category Y, and no object from category Y is available, we can use an object C from category Z to occlude the same portion of A. This results in a similar occlusion effect, demonstrating that the specific category of the occluding object is less important than replicating the occlusion relationship.

\begin{figure}[t]                
	\centering                      
	\includegraphics[height=5.5cm,width=7.5cm]{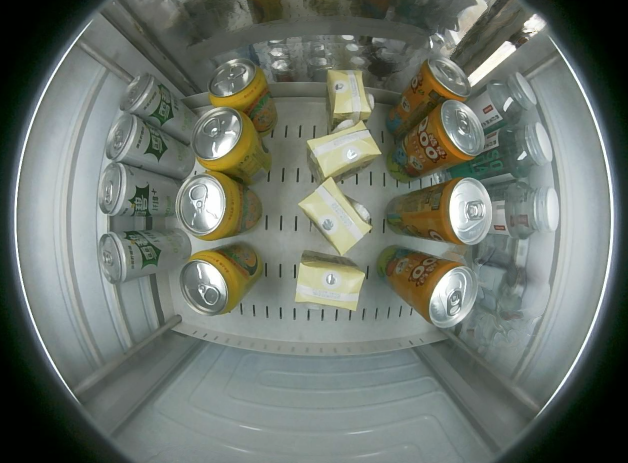}  
	\caption{\centering{Example of a beverage-only arrangement.}}      
	\label{Fig1}                     
\end{figure}

\begin{figure}[t]                  
	\centering                      
	\includegraphics[height=5.5cm,width=7.5cm]{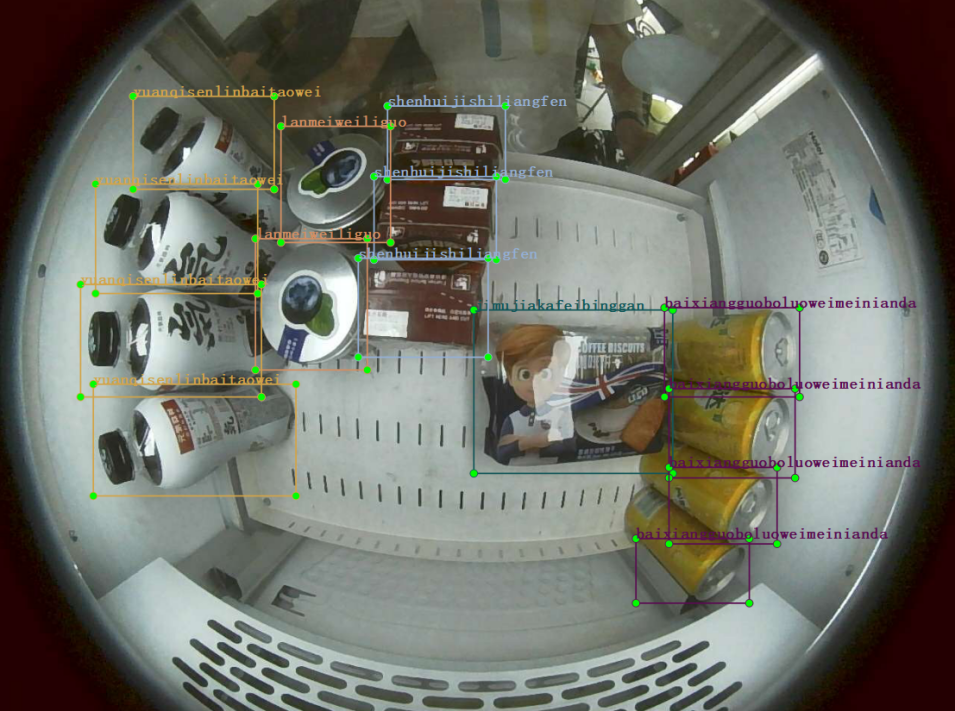}  
	\caption{\centering{Example of a beverage and snack arrangement.}}      
	\label{Fig2}                     
\end{figure}

We illustrate this concept with the typical placement of goods on a shelf, as shown in Figures \ref{Fig1} and \ref{Fig2}. Since we use a fisheye camera with an ultra-wide-angle lens that introduces strong visual distortion to create hemispherical images, we must arrange the items carefully to ensure all goods are visible in the camera’s view. Tall items, such as beverages, should be placed near the shelf wall, while shorter items are positioned centrally. As the size of items increases, they should be placed more peripherally. This ensures that all objects remain visible to the fisheye camera and can be detected by the neural network model.

Preparing objects of various sizes is crucial for simulating a wide range of occlusion relationships. For example, in real-world scenarios, a target object may be partially occluded by smaller objects, while larger objects can obscure more significant portions of the target.

Occlusions can be introduced during either the data collection or data augmentation stages. During data collection, the size of each object category plays a critical role in generating realistic occlusion relationships, as different object sizes naturally produce different types of occlusions. In contrast, during the data augmentation stage, object size is less critical, as we can use any category to occlude a target object. Techniques such as copy-paste, cut-paste, image scaling, and image translation can be employed to simulate these occlusions effectively.

\subsection{Occlusion relationship}

Accurately identifying the real-world object occlusion distribution is essential for effective imitation. In specific scenarios, object occlusion depends on various factors, including camera viewpoint, object size, and object placement. Object relationships must be reasonable; for example, in indoor settings, a cup on a table might be partially or fully occluded by a paper picker depending on the viewpoint, while a TV remote is more likely to be placed beside the cup rather than on top of it. Therefore, we prioritize collecting common occlusion relationships, ensuring that each type of occlusion is represented by one or two cases, which is sufficient to achieve high accuracy in real-world test cases.

To simulate this occlusion distribution, we apply a Monte Carlo method to sample data points. First, we identify the occlusion distribution of the new category in real scenarios and process each category individually. Then, we generate synthetic images by applying the copy-paste technique to occlude objects from the new category according to this distribution. These synthetic images are paired with a few real images to train the detection network.

During the data collection stage, we lack the specific occlusion distribution for a new category, but we may have access to a similar-sized category from previous data. When generating occlusions, the surface material or texture of the new category is not crucial. However, the placement of the new category depends on its size and intrinsic characteristics. For example, in the FVSS dataset, smaller items like packaged snacks or canned drinks are placed in the center of a shelf layer, while larger items, such as snack bags, are positioned on the periphery.

To maximize space utilization on the shelf, goods are arranged to ensure they are all visible from the top-centered fisheye camera, with each item’s visible region distinct enough for human recognition. The top or top-lateral part of each item must be visible, avoiding any stacking of goods. In a fully packed layer, the lower parts of objects are typically occluded by adjacent items. Smaller items are placed centrally, while larger ones are positioned on the edges of the layer or near the shelf walls. This arrangement minimizes occlusion by items of the same or different categories.

For example, when adding a new large item, such as a water bottle, it would typically be placed on the periphery or near the shelf wall to reduce occlusion by nearby objects. The occlusion ratio varies depending on the adjacent objects: a bottle of water may have only its cap visible if occluded by an identical bottle, while a smaller milk carton could obscure up to two-thirds of the bottle. A lying-down snack bag might only occlude about one-third. Additionally, placing larger objects near the fisheye camera center should be avoided to prevent total occlusion of smaller items.

In the data-occlusion stage, we generate new occluded images by using already annotated images, following the occlusion distribution identified for the target new category. The first step is determining the correct occlusion distribution for the new category. The most reasonable approach involves analyzing the new category's attributes and inferring the occlusion distribution based on the expertise of experienced researchers. However, this method is difficult to generalize, as it requires expert knowledge for each new category. Therefore, an automated approach is needed.

For example, in the COCO dataset, the 'person' category is often annotated in scenarios such as standing on the street or sitting at a table. A person may be occluded by other people outdoors or by a table indoors. Interestingly, the head of the person is almost always visible, even in crowded or distant scenes. If a person's head is not visible, it likely means the dataset organizers did not collect such images, as humans typically recognize others by their heads. Additionally, annotators may be reluctant to label an image where only the lower half of a person’s body is visible, with no head in sight.

Some notable features for occlusion in object detection are as follows: 1) A person's head may be occluded by an umbrella. 2) The head may appear in a lateral view or show the back side, which should be considered during data collection. In the data augmentation stage, the focus should be on imitating the real occlusion relationships, not the varying viewpoints. 3) In rare cases, an image might show only a small part of a human, such as close-up hands or a foot, while still being annotated as a person. These features can be generalized to other categories, including animals, which typically present their heads in pictures to facilitate recognition by the observer.

The COCO dataset, being highly diverse, includes numerous environments, lighting conditions, gestures, and viewpoints. For instance, the 'bear' category has multiple sub-categories like polar bear, black bear, brown bear, and raccoon. By adding a new category with only a small number of images—such as dozens of images—it is possible to train a detection network effectively, incorporating both the new and existing categories.

For smaller objects, like toothbrushes or remote controllers, the objects may be captured in both close-up and distant views, leading to significant variation in object sizes within images. A common question arises: is it useful to apply small object occlusion distributions to larger objects? Our findings suggest that it is indeed useful, and performance can be further improved by applying image scaling as a data augmentation technique.

We also analyzed the Open Images Dataset \cite{kuznetsova2020open}, which contains a larger number of samples and greater diversity across categories. The dataset provides four types of annotations: Detection, Segmentation, Relationships, and Localized Narratives. Detection annotations use bounding boxes, while Segmentation annotations are represented by polygons. The Relationships annotation captures various interactions between humans and objects or between different objects, with dotted-line bounding boxes indicating one object being contained within another. These relationships are closely tied to the data occlusion distribution of categories and offer valuable insights for our work. 

\subsection{Camera viewpoints}
Imitating all possible viewpoints, particularly in large outdoor scenarios, can be challenging. However, we found a simple solution that achieves high accuracy in real-world settings. We adopted the approach from NERFIES \cite{Park_2021_ICCV}, using the main camera viewpoint along with several slightly offset viewpoints to capture images of objects, while ignoring rare or extreme viewpoints.
\subsection{Copy-paste augmentation}

After collecting dozens of images for our new SKUs, we employ the copy-paste data augmentation strategy \cite{ghiasi2021simple} to cover more sample points representing data placement and occlusion distributions, thereby improving detection performance. The copy-paste strategy involves randomly transferring bounding box regions from one image to another according to our data occlusion distribution, while ensuring minimal overlap with existing objects.

\subsection{FairMOT-based annotation}
bounding boxes are primarily used for annotating objects. In controlled data collection scenarios, we can move objects slowly across frames, enabling the use of tracking models to annotate each object in continuous movement, thereby reducing the workload of human annotators.

We initially tested single object tracking (SOT). By slowly moving one object in a clip, we can annotate the target object in the first frame while keeping others static. We assume the camera remains stationary, though moving the camera slowly can enhance SOT performance. However, two issues arise with SOT: 1) When multiple objects of the same category are placed near each other, the tracker may shift to a nearby object, requiring a more accurate SOT model for scalable annotation. 2) Since SOT supports only single-object tracking, many clips are needed to annotate different objects individually. To address these issues, we adopt multi-object tracking (MOT) using FairMOT, enabling simultaneous tracking of multiple objects and further streamlining the annotation process.

\begin{figure}[t]                  
	\centering                      
	\includegraphics[height=5.5cm,width=7.5cm]{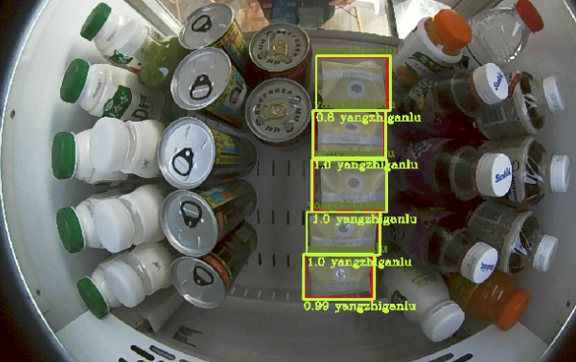}  
	\caption{\centering{Detection results for small box-shaped drinks.}}      
	\label{Fig3}
\end{figure}

\begin{figure}[t]                   
	\centering                      
	\includegraphics[height=5.5cm,width=7.5cm]{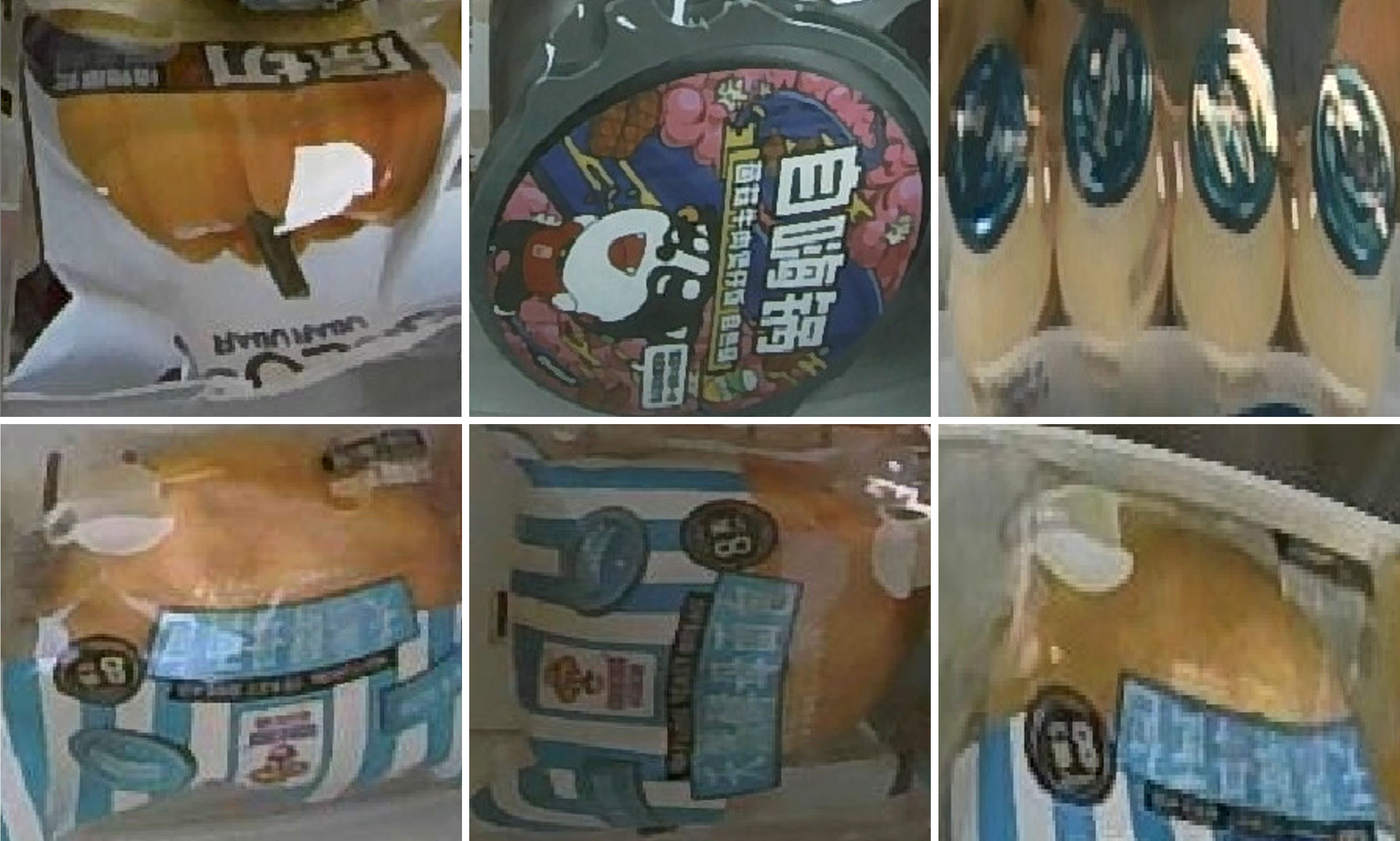}  
	\caption{\centering{Examples of low-height snacks.}}      
	\label{Fig4}
\end{figure} 

\section{Experiments}

Experiments are conducted to demonstrate that our approach requires only tens of images for each new category to achieve comparable accuracy to models trained with thousands of images. We explore two experimental directions: data occlusion in the data collection stage and data occlusion in the data augmentation stage.

Our dataset, Fisheye View of Shelf SKUs (FVSS), is used for validation in the data collection stage. The dataset provides fisheye camera views of a shelf layer, as shown in Fig. \ref{Fig3} and Fig. \ref{Fig4}, with bounding boxes annotated for each category. In these experiments, we use hundreds of categories as the base dataset and attempt to add a new category.

For the data augmentation stage, we use the COCO dataset as our testbed. COCO consists of 80 categories, from which we randomly select one new category and use the remaining 79 as the base dataset. We analyze the occlusion distribution for the new category and select 1\% to 10\% of images that represent key sample points for training. For detection, we use the YOLOv5-small model and convert all annotations to the YOLOv5 format.

\subsection{Data collection}

\begin{figure}[h]                  
	\centering                      
	\includegraphics[height=3.5cm,width=7.5cm]{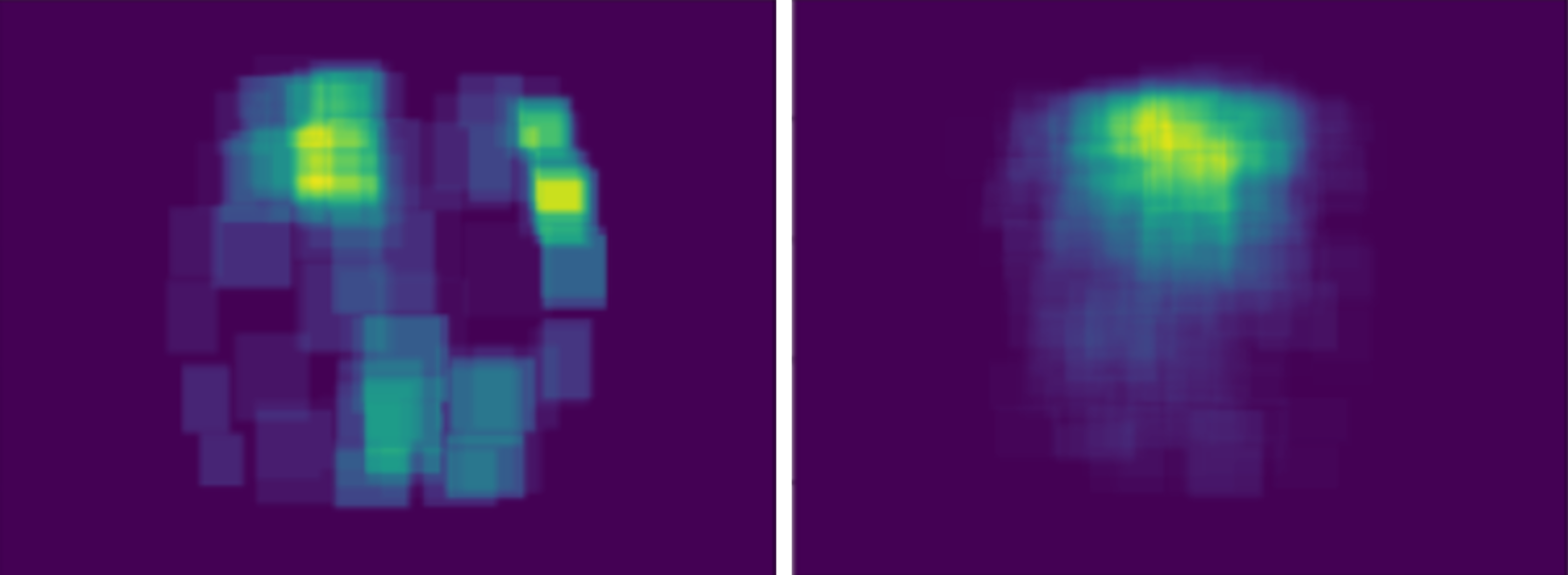}  
	\caption{\centering{Heatmaps of two categories, "guangshiboluopi" (left), and "yangzhiganlu" (right).} }    
	\label{Fig5}                     
\end{figure}

Experiments are conducted in a shelf environment using fisheye cameras, following the FVSS dataset construction style. Our base training dataset consists of 10,000 images covering 457 categories. We add one new category with only 10 images to the base dataset and evaluate performance on a validation set of 1,000 images, each containing at least one bounding box for the new category. The heatmaps for two categories in our dataset, relevant to their occlusion distribution, are shown in Fig. \ref{Fig5}.

For example, 10 images of "coke can" are added as a new category to a training dataset of 10,000 images containing 457 categories, none of which include "coke can." These 10 images are carefully selected from important sample points within the data occlusion distribution for "coke can" in a shelf environment, resulting in 58 bounding boxes. We then construct a validation dataset of 1,000 images, covering 179 categories, with each image containing at least one "coke can" bounding box (totaling 3,939 bounding boxes). We evaluate performance using three metrics: 1) AP@0.5 and AP@0.5:0.95 for "coke can" in the validation dataset; 2) pass rate measures if all "coke can" instances are correctly detected; and 3) wrong-class rate indicates whether a "coke can" is mistakenly classified as another category with confidence lower than 95\%. Results are shown in Table \ref{table1}.

\begin{table*}[htbp]
	\normalsize
	\centering
	\caption{Results specific to "coke can".}
	\begin{tabular}{ccccc}\hline
		AP@0.5&AP@0.5:0.95&pass rate&wrong-class rate & wrong-class rate@0.95 \\\hline 
		98.4\% & 83.6\% & 81.3\% & 77.0\% & 87.0\%\\\hline
	\end{tabular}
\label{table1}
\end{table*}

16 different categories are tested, each added as a new category in isolation. The categories are all retail field goods, such as snacks, milk, beverages, and more. Our findings conclude that we can train a new category with just a few images while maintaining an accuracy above 80\%, and a wrong-class rate above 85\% on average. This means that only 3\% of images in the validation dataset are either wrong-classed with high confidence or ignored by the detection model. Some category results are shown in Table \ref{table2}. The wrong-class rate is defined as instances where confidence is below 90\%. We also introduce a new metric, the "severe error rate," which specifically measures the rate at which bounding boxes are miss-identified as another category with a confidence above 90\% or are not detected at all. Results are presented in Table \ref{table2} and Table \ref{table3}.

Additionally, we find that using a wide variety of different-sized categories to generate diverse data occlusion relationships significantly improves the model's performance, nearly doubling the average accuracy.

\begin{table*}[t]      
	\centering
	\caption{Experiments regarding each class as a new category.}
	\begin{tabular}{ccccc}\hline
		name & image number & pass rate & wrong-class rate & severe error rate \\\hline
		xiandangao & 6015 & 78\% & 91\% & 1.98\% \\
		yibaochunjingshui & 2037  & 54\% & 92\% & 3.68\% \\
		jiaduobaoguan & 1238   & 42\% & 78\% & 12.76\% \\
		cuiguoba & 6359   & 91\% & 80\% & 1.8\% \\
		420meizhiyuanguolicheng & 712    & 95\% & 95\% & 0.25\% \\
		feizixiaolizhi & 1884    & 52\% & 95\% & 2.4\% \\
		heqingjiaotangbinggan & 3569    & 84\% & 92\% & 1.28\% \\
		duoweixiaoxibing200 & 1799    & 90\% & 90\% & 1.0\% \\
		4wahahaadgainai & 2807    & 55\% & 100\% & 0.0\% \\
		yizhongtaohuangtaoguantou & 2520    & 78\% & 80\% & 4.4\% \\
		heqingjiaotangbinggan & 3992     & 94\% & 86\% & 0.84\% \\
		4wahahaadgainai & 3094     & 32\% & 91\% & 6.21\% \\
		enaakdianxinmian30g & 488     & 62\% & 76\% & 9.12\% \\
		guowangshiguangguoba & 867     & 90\% & 100\% & 0.0\% \\
		mailisu & 531     & 95\% & 95\% & 0.25\% \\
		average & 3194 & 72\% & 90.7\% & 4.2\%\\\hline
	\end{tabular}
\label{table2}
\end{table*}

\begin{table*}[t]
	\centering
	\caption{Zero-shot vs Few-shot.}
	\begin{tabular}{ccc}\hline
		name & pass rate & wrong-class rate \\\hline
		w/o new category data & 0.0\% & 79.0\% \\
		w new category data & 72.0\% & 90.0\% \\\hline
	\end{tabular}
	\label{table3}
\end{table*}

\begin{table*}[h]
	\centering
	\caption{Zero-shot results.}
	\begin{tabular}{ccc}\hline
		name & image number & wrong-classed rate \\\hline
		xiandangao & 6015 & 87.0\% \\
		yibaochunjingshui & 2030 & 87.0\% \\
		jiaduobaoguan &  2945 & 81.0\% \\
		cuiguoba &  6992 & 90.0\% \\
		420meizhiyuanguolicheng &  712 & 78.0\% \\
		feizixiaolizhi &  1884 & 87.0\% \\
		heqingjiaotangbinggan &  3569 & 52.0\% \\
		duoweixiaoxibing200 &  1799 & 93.0\% \\
		4wahahaadgainai &  2805  & 44.0\% \\
		yizhongtaohuangtaoguantou &  2520  & 82.0\% \\
		average &  3127.1  & 79.3\% \\\hline
	\end{tabular}
	\label{table4}
\end{table*}

\begin{table*}[htbp]
	\centering
	\caption{Comparison for new category training.}
	\begin{tabular}{cccc}\hline
		sku name & 3000+ bboxes & 60 bboxes (20 images) & 370 bboxes (20 images + copy-paste)\\\hline
		guangshiboluopi & 33.83\% & 12.7\% & 53.38\%\\
		yangzhiganlu & 60.96\% & 34.76\% & 76.83\%\\
		zhiqingchunniunai & 49.87\% & 3.56\% & 27.95\%\\
		tengyeyicunxiaoyuanbinggan & 95.98\% & 38.16\% & 98.19\%\\
		aolangtangeweihuabinggan & 37.50\% & 58.33\% & 97.22\%\\
		average & 55.63\% & 29.52\% & 70.71\%\\\hline
	\end{tabular}
	\label{table5}
\end{table*}

With adding images of a new category from a different domain, such as images captured by hand-held smartphone, cross-domain experiments are conducted, alongside shelf fisheye images. No domain adaptation methods were applied for enhancement. The results showed a 0\% pass rate for the new category in a validation dataset of 1000 images. The wrong-class rate for the new category was nearly identical to the case where no new category data was added. This highlights that domain adaptation remains a challenge when training with new categories. The results are shown in Table \ref{table3}.

Next, the effect of adding a new category to the training dataset is assesed. In the validation dataset, which contained new category data, we discovered that if the new category was not included in the training dataset, the category could not be detected correctly in any of the validation images, resulting in a 0\% pass rate. Additionally, 21\% of the bounding boxes were either missed or incorrectly detected as other categories with high confidence. However, when we added only 10 images of the new category to the training dataset, the pass rate for the new category in the validation dataset increased to 72\%, and the wrong-class rate at 0.90 confidence decreased to 90\%. We also observed that categories with a high aspect ratio tended to show a larger increase in wrong-class rate at 0.90 confidence. Adding images of these high aspect ratio categories may reduce the chances of wrong-classs as other categories. Despite this, high aspect ratio categories tend to have lower pass rates when trained with only a few images. This suggests that even a small number of images can significantly improve the detection of a new category, especially when those images fit well into the occlusion distribution. The neural network can then learn the most necessary and important features. The results are shown in Table \ref{table4}.

\begin{figure}                   
	\centering                      
	\includegraphics[height=4.5cm,width=7.5cm]{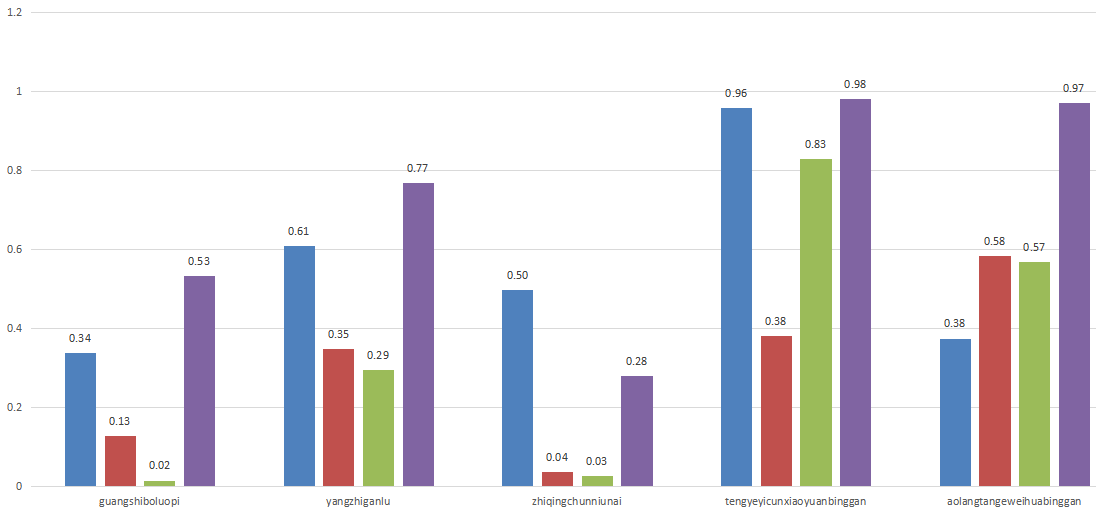}
	\caption{\centering{Comparison of copy-paste data augmentation results: Blue represents the model trained with the original 3000+ bounding boxes of the target class. Red shows the performance using only 15 randomly sampled bounding boxes. Green indicates results from training with 17 close-view smartphone-captured images taken from various angles. Purple shows the combined result using both the 17 close-view smartphone images and 370 randomly sampled bounding boxes.} }
	\label{Fig6}                     
\end{figure}

\begin{figure}                   
	\centering                      
	\includegraphics[height=4.5cm,width=7.5cm]{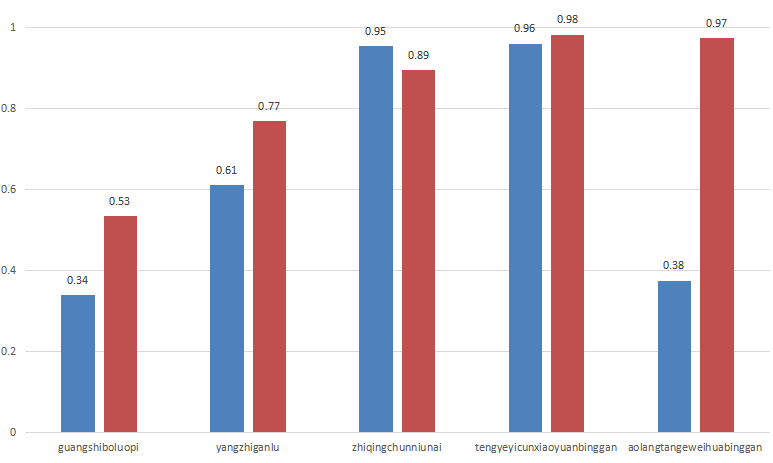}
	\caption{\centering{Results of copy-pasted "zhiqingchunniunai". Blue represents the model trained with the original 3000+ bounding boxes of the target class. Red shows the combined result using both the 17 close-view smartphone images and 370 randomly sampled bounding boxes.}}
	\label{Fig7}                     
\end{figure}

In a further experiment, we tested adding a new category with just a few images to a large dataset. We used a large shelf fisheye view dataset containing over 360,000 images and added a new category, "sizhoushaokaoweixiatiao," with only 15 images and 60 bounding boxes. After training for 1.5 epochs with common data augmentation techniques such as image flipping, hue tuning, and normalization, we evaluated the model on a test dataset of 500 real images. Only about 10 images were incorrectly classified. This result demonstrates the potential of using the copy-paste strategy along with data occlusion distribution to train effective detection models using bounding box annotations. The results are shown in Figures \ref{Fig6} and \ref{Fig7}.

\subsection{Data augmentation}
the comparison is using the training with a normal dataset containing more than 3,000 bounding boxes to training with only 20 new category images. These 20 images are a subset of the large dataset mentioned earlier. We performed two types of experiments with the 20 new category images.

\begin{figure}                   
	\centering                      
	\includegraphics[height=6.5cm,width=9.5cm]{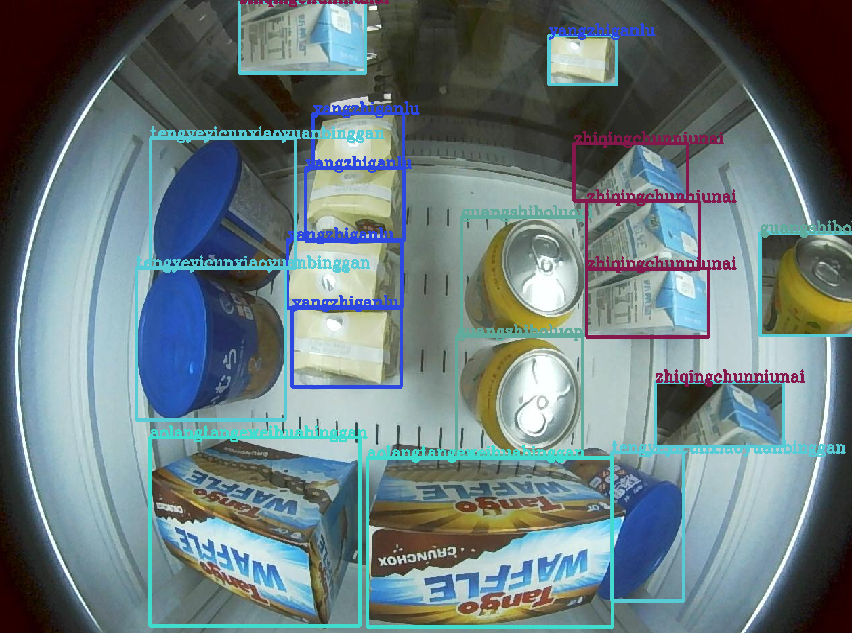}
	\caption{\centering{Example of copy-pasted bounding boxes.}}
	\label{Fig8}                     
\end{figure}  

\begin{figure*}
	\centering
	\includegraphics[height=7.0cm,width=17.5cm]{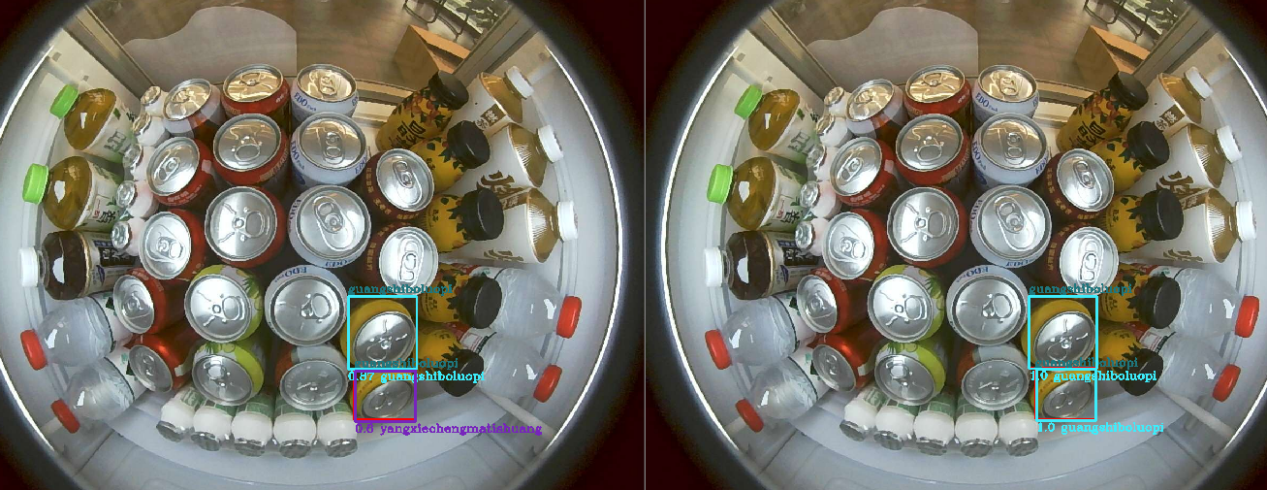}
	\caption{\centering{Example of a detection failure case. The two categories exhibit high visual similarity.}}
	\label{Fig9}                     
\end{figure*}

In the first experiment, we trained the model using only these 20 images, applying data augmentation techniques such as image flipping, HSV transformation, and hue tuning. In the second experiment, we used the copy-paste data augmentation strategy following the data occlusion distribution to generate an additional 100 images from the original 20 images. This brought the total number of new category images to 120, with 100 of them being copy-pasted. We tested the model with five new categories, one at a time, and the results are shown in Table \ref{table5}. The results were remarkable, showing that using a small number of images combined with copy-paste augmentation outperformed training on the original large dataset. Figure \ref{Fig8} shows an image augmented using the copy-paste strategy we employed, while Figure \ref{Fig9} illustrates a failure case of detection by a network trained using our approach.

In certain situations, we may use an already collected dataset for training, and we cannot control the data collection stage. However, we still want to add a few images of a new category to the existing dataset. To demonstrate that even a small number of images of a new category can achieve relatively high accuracy, we designed experiments where we select these images from the important sample points of the data occlusion distribution for the new category in the test dataset.

\begin{figure*}                   
	\centering                      
	\includegraphics[height=8.5cm,width=16.5cm]{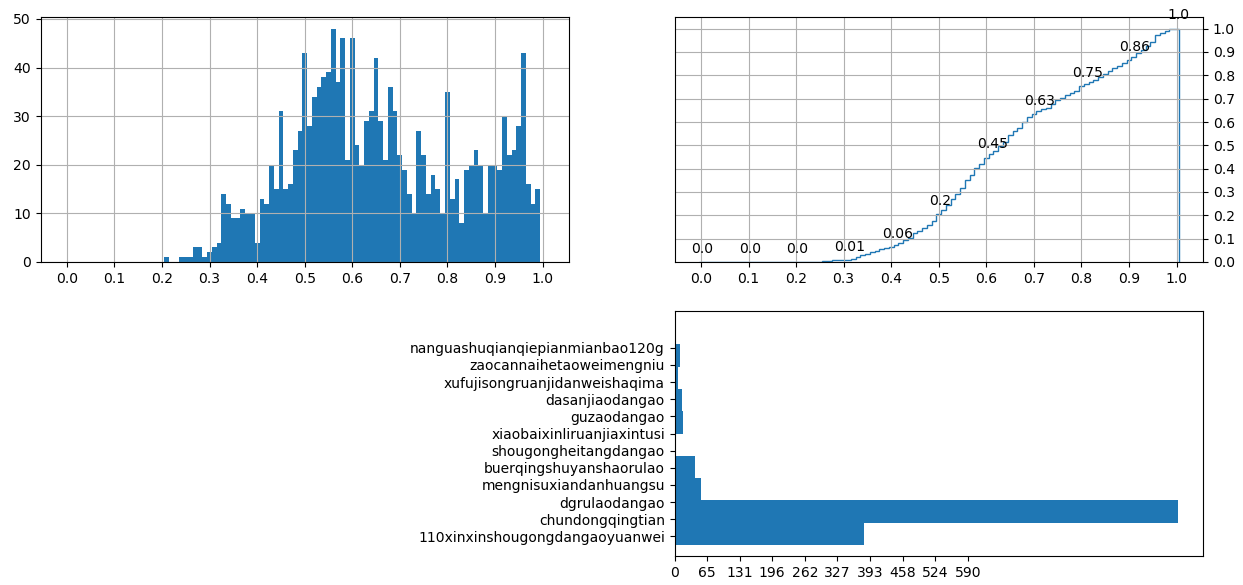}  
	\caption{\centering{Detection results based on 30 images of "xiandangao", with pass rate 79\%. The figure includes three plots: confidence distribution (top-left), accumulated confidence (top-right), and the number of detected classes (bottom-right).}}
	\label{Fig10}                     
\end{figure*}

\begin{figure*}                   
	\centering                      
	\includegraphics[height=8.5cm,width=16.5cm]{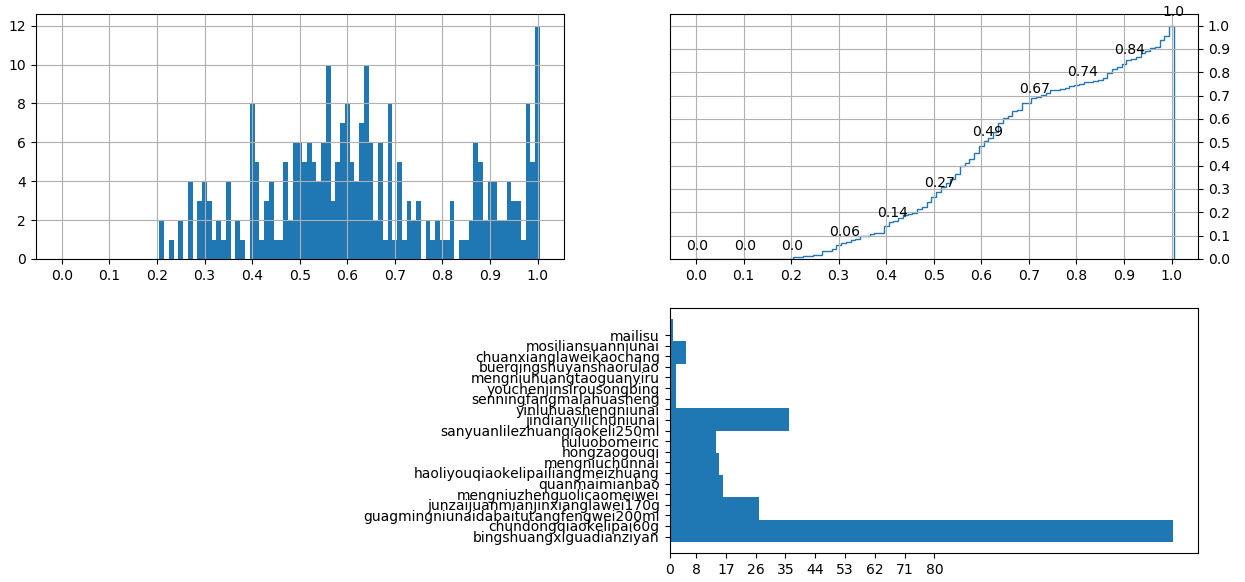}  
	\caption{\centering{Detection results based on 30 images of "heqingjiaotangbinggan", with pass rate 93\%. There are three sub-plots, including confidence distribution (top-left), accumulated confidence (top-right), and the number of detected classes (bottom-right).}}
	\label{Fig11}                     
\end{figure*}

Our implementation is inspired by the approach in \cite{ghiasi2021simple}, where they utilize a simple copy-paste data augmentation strategy to achieve noticeable improvements in accuracy. We believe that this conclusion is driven by the fact that the copy-paste operations generate many new occlusion relationships, capturing important sample points of the data occlusion distribution. Figures \ref{Fig10} and \ref{Fig11} show the test results for two categories, which illustrate the confidence distribution of the target categories in the test dataset.

\section{Conclusion}
Data collection is a core step when applying vision systems to real-world tasks. In this paper, we propose an object occlusion data collection method, which has proven to be both effective and robust. Object occlusion performs well across multiple experimental settings and leads to significant improvements, even with a small amount of data. Our experiments are based on the FVSS dataset and COCO benchmarks.

The object occlusion data collection and augmentation strategy we propose is simple to integrate into any dataset, whether constructing a new dataset or adding new categories to an existing one. This approach reduces training costs by requiring only a small number of images. Consequently, we can use smaller models with suitable data occlusion strategies—such as the copy-paste technique—to create appropriate occlusion relationships for target objects. This method also uses less memory during the training process. Proper object occlusion data collection and augmentation strategies allow small models to achieve accuracy comparable to more complex models.

Our findings show that networks can learn a new category from only a few samples, similar to how humans, with their strong inference abilities, learn. On the other hand, human learning also requires mimicking network learning styles, which involves minimal analysis and inference ability but exposure to more samples. This suggests that the learning process of networks, which typically involves presenting more examples without detailed explanations, could be beneficial for learning new concepts or languages. Future work could focus on improving object occlusion data collection and augmentation strategies for more type of objects.

\bibliography{manuscript}
\end{document}